\documentclass[final]{cvpr}

\usepackage{times}
\usepackage{epsfig}
\usepackage{graphicx}
\usepackage{amsmath}
\usepackage{amssymb}
\usepackage{booktabs}       %
\usepackage{comment}
\usepackage{subcaption}

\usepackage[pagebackref=true,breaklinks=true,colorlinks,bookmarks=false]{hyperref}

\def\cvprPaperID{4796} %
\def\confYear{CVPR 2021}

\DeclareFontFamily{OT1}{pzc}{}
\DeclareFontShape{OT1}{pzc}{m}{it}{<-> s * [1.10] pzcmi7t}{}
\DeclareMathAlphabet{\mathpzc}{OT1}{pzc}{m}{it}

\begin{document}

\newcommand{\OURS}{Neural Deformation Graph}

\definecolor{abcolor}{RGB}{150,150,50}
\newcommand\AB[1] {\emph{\textcolor{abcolor}{AB: #1}}}

\definecolor{mzcolor}{RGB}{0,0,255}
\newcommand\MZ[1] {\emph{\textcolor{mzcolor}{MZ: #1}}}

\definecolor{jtcolor}{RGB}{255,0,255}
\newcommand\JT[1] {\emph{\textcolor{jtcolor}{JT: #1}}}

\definecolor{adcolor}{RGB}{50,150,150}
\newcommand\AD[1] {\emph{\textcolor{adcolor}{AD: #1}}}

\newcommand{\MATTHIAS}[1]{{\bf\textcolor{red}{Matthias: #1}}}

\definecolor{ppcolor}{RGB}{50,150,0}
\newcommand{\PP}[1]{{\bf\textcolor{ppcolor}{PP: #1}}}

\newcommand{\TODO}[1]{{\textcolor{red}{TODO: #1}}}
\newcommand{\rev}[1]{{\textcolor{blue}{#1}}}

\title{Neural Deformation Graphs for Globally-consistent Non-rigid Reconstruction}

\author{
Aljaž Božič$^1$~~~
Pablo Palafox$^1$~~~
Michael Zollhöfer$^2$~~~
Justus Thies$^1$~~~
Angela Dai$^1$~~~
Matthias Nie{\ss}ner$^1$
\vspace{0.2cm} \\ 
$^1$Technical University of Munich~~~
$^2$Facebook Reality Labs
\vspace{0.2cm} \\ 
}

\twocolumn[{
	\renewcommand\twocolumn[1][]{#1}%
	\maketitle
	\begin{center}
		\includegraphics[width=1.0\textwidth]{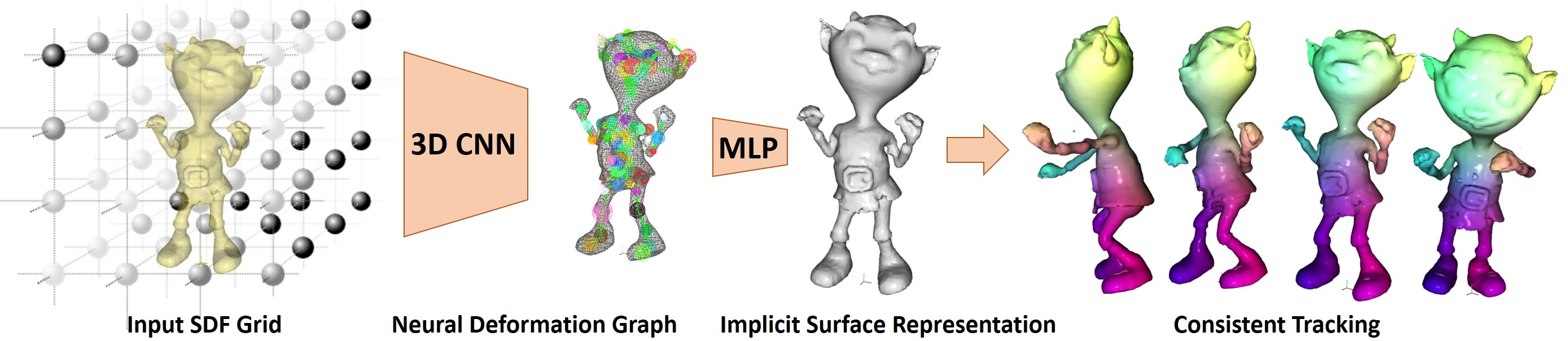}
		\captionof{figure}{Neural Deformation Graphs: given range input data, represented as a signed distance field, our method predicts globally-consistent deformation graph that is used to reconstruct the non-rigidly deforming surface of an object. The surface of the object is represented as a set of implicit functions centered around the deformation graph nodes. Our global optimization provides consistent surface and deformation prediction, enabling robust tracking of an observed input sequence and even multiple disjoint captures of the same object (as we do not assume sequential input data).}
		\label{fig:teaser}
	\end{center}
}]

\begin{abstract}
We introduce \OURS s for globally-consistent deformation tracking and 3D reconstruction of non-rigid objects.
Specifically, we implicitly model a deformation graph via a deep neural network.
This neural deformation graph does not rely on any object-specific structure and, thus, can be applied to general non-rigid deformation tracking.
Our method globally optimizes this neural graph on a given sequence of depth camera observations of a non-rigidly moving object.
Based on explicit viewpoint consistency as well as inter-frame graph and surface consistency constraints,  the underlying network is trained in a self-supervised fashion.
We additionally optimize for the geometry of the object with an implicit deformable multi-MLP shape representation.
Our approach does not assume sequential input data, thus enabling robust tracking of fast motions or even temporally disconnected recordings.
Our experiments demonstrate that our \OURS s outperform state-of-the-art non-rigid reconstruction approaches both qualitatively and quantitatively, with $64$\% improved reconstruction and $62$\% improved deformation tracking performance.
\end{abstract}

\section{Introduction}

Capturing non-rigidly deforming surfaces is essential towards reconstructing and understanding real-world environments, which are often highly dynamic.
While impressive advances have been made in reconstructing static 3D scenes~\cite{curless1996volumetric,kazhdan2006poisson}, dynamic tracking and reconstruction remains very challenging. 
A plethora of domain-specific dynamic tracking methods has been developed (e.g., human bodies, faces, hands), leveraging strong domain shape and motion priors for robust tracking~\cite{cao2018openpose, loper2015smpl, mueller2018ganeratedhands, thies2016face2face}.
However, real-world environments encompass a vast diversity of deformable objects -- including people with clothing or animals -- making domain specific shape priors often intractable for general deformable reconstruction; in this work, we thus focus on general non-rigid 3D reconstruction without shape or motion priors for general object tracking and reconstruction.

A seminal work in non-rigid 3D reconstruction is DynamicFusion~\cite{newcombe2015dynamicfusion}, which was the first approach to demonstrate real-time dense reconstruction of dynamic scenes using just a single RGB-D sensor.
DynamicFusion showed promissing results towards dynamic reconstruction, but still struggles in many real-world scenarios, which typically include strong deformations and fast frame-to-frame motion, due to its low-level, local correspondence association step.
In particular, the incremental construction of a deformation graph is prone to error aggregation and can lead to tracking failures.
Recently, data-driven methods based on deep learning have been introduced~\cite{bozic2020deepdeform, li2020learning, bozic2020neuraltracking} that learn priors of non-rigidly deforming objects from dense flow annotations. %
These approaches leverage a similar incremental deformation graph construction as DynamicFusion, but learn to establish more robust tracking via more sophisticated correspondence optimization based on data-driven priors.
However, despite more robust correspondences, these methods still operate on a frame-by-frame basis, thus, aggregate tracking errors and are unable to recover if tracking fails.
In order to address these shortcomings without assuming data-driven priors, we propose a globally-consistent neural deformation graph which allows for non-rigid reconstruction from commodity sensor observations, represented as signed distance fields (see Fig.~\ref{fig:teaser}).
The neural deformation graph gives access to the per frame deformation graph nodes and stores the global graph connectivity.
To robustly optimize for consistent deformations over fast motions, we introduce viewpoint consistency (independently for every frame)  as well as  graph and surface consistency constraints (between pairs of frames).
Our viewpoint consistency loss measures the consistency of graph node position predictions w.r.t. rotation augmentation.
The graph and surface consistency losses encourage deformations to be modeled in our \OURS{} such that local graph edge distances are preserved between frames and the deformed surface geometry of a source frame aligns well with the geometry of the target frame.
Additionally, our approach does not assume temporally close frames, thus making it easily applicable to low FPS settings or the combination of independently captured depth recordings.

Since there exists no general canonical pose (like a T-pose of a human~\cite{loper2015smpl}) that fits all deformable objects, we avoid modeling it explicitly.
Instead, we propose to employ a set of implicit functions that are centered around the deformation graph nodes.
Specifically, we model local signed distance functions (SDFs) using multi-layer perceptrons (MLPs) that can be deformed to fit any frame, without requiring an explicit canonical pose.
The global shape is evaluated by the integration of these local MLP predictions.

\vspace{0.25cm}
\noindent
To summarize, our technical contributions are: 
\begin{itemize}
    \vspace{-0.1cm}
    \item a globally-optimized deformation graph that is able to handle deformations present in all frames of an unstructured dataset or a sequence of an object;
    \vspace{-0.1cm}
    \item a combination of per-frame viewpoint consistency and frame-to-frame graph and surface consistency for robust tracking of fast deformations;
    \vspace{-0.1cm}
    \item an implicit deformable multi-MLP shape representation anchored on the scene-specific deformation graph.
\end{itemize}

\section{Related Work}
Our approach is leveraging a low dimensional deformation graph to model the non-rigid deformations of an object, while the actual surface is represented by an implicit function by means of a multi-layer perceptron (MLP).
We will discuss the most related approaches in these two fields. 

\paragraph{Non-rigid Reconstruction}
Non-rigid reconstruction is a highly active research field, in particular using commodity RGB-D sensors such as the Kinect.
The seminal work DynamicFusion of Newcombe et al. \cite{newcombe2015dynamicfusion} tracks deformable motion and reconstructs the object's shape in an incremental fashion, i.e., frame-by-frame.
While this approach relies on local depth correspondences, follow-up methods additionally use sparse SIFT features \cite{innmann2016volumedeform}, dense color tracking \cite{guo2017monofvv} or dense SDF alignment \cite{slavcheva2017killingfusion, slavcheva2018sobolevfusion}.
These methods show impressive results, but often struggle with fast frame-to-frame motion given their use of hand-crafted correspondences.
Bozic et al.~\cite{bozic2020deepdeform} introduced an annotated dataset of non-rigid motions that allows to train data-driven non-rigid reconstruction methods with learned correspondences~\cite{bozic2020deepdeform, li2020learning, bozic2020neuraltracking}.
While learned correspondences improve tracking performance, the approaches are still inherently limited by the employed frame-to-frame tracking paradigm, i.e., tracking errors accumulate over time, and if tracking is lost it is unable to recover.
Tracking robustness can also be improved without any learned priors by using multi-view input data~\cite{collet2015fvv,guo2019relightables} (setups with more than $50$ cameras) and high-speed cameras~\cite{dou2017motion2fusion} ($8$ cameras at $200$ frames per second (FPS)).
In contrast to these frame-to-frame tracking approaches, there are methods that focus on global non-rigid optimization \cite{dou20153d, wang2018dynamic, innmann2020nrmvs, sidhu2020neuralnonrigidsfm}; however, these methods either assume ground-truth optical flow \cite{sidhu2020neuralnonrigidsfm}, or they share the same drawbacks of the aforementioned frame-to-frame tracking approaches \cite{dou20153d, wang2018dynamic, innmann2020nrmvs}, and thus have difficulties handling fast deformable motion.

\begin{figure*}
\includegraphics[width=\linewidth]{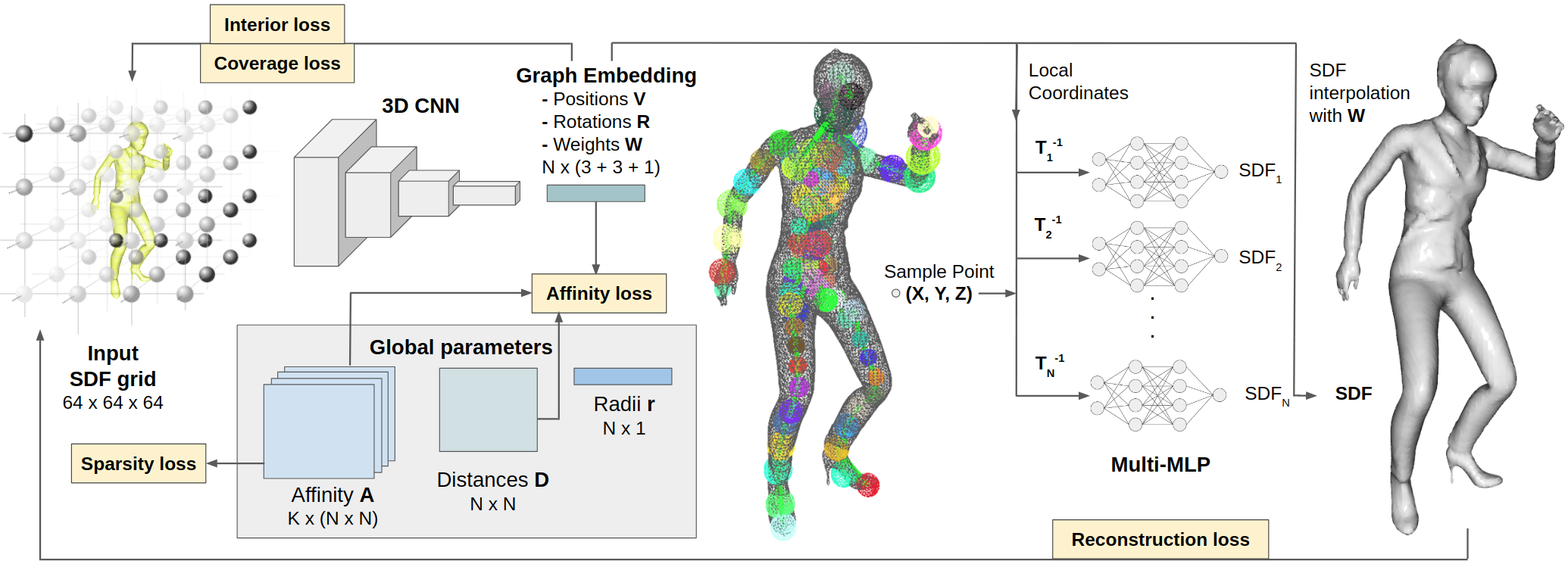}
\caption{
A Neural Deformation Graph encodes a $64^3$ SDF grid to a graph embedding with graph node positions $\mathbf{V}$, rotations $\mathbf{R}$ and importance weights $\mathbf{W}$.
To compute an SDF value for a sample point $(X, Y, Z) \in \mathbb{R}^3$, the point is transformed to local coordinates around each node, and passed through locally embedded implicit functions that are represented as MLPs; the global SDF value is computed by interpolating the local MLP predictions using the node radii $\mathbf{r}$ and importance weights $\mathbf{W}$.
For graph regularization, a set of affinity matrices $\mathbf{A}_i \in \mathbb{R}^{N \times N}$ and a node-to-node distance matrix $\mathbf{D} \in \mathbb{R}^{N \times N}$ are globally optimized.
}
\label{fig:method-overview}
\end{figure*}

\paragraph{Deformation Graphs}
State-of-the-art non-rigid reconstruction methods often model deformations with a sparse deformation graph, following the Embedded Deformation \cite{sumner2007ed} formulation.
Deformation graphs offer a robust alternative to dense motion estimation with optical flow or scene flow methods, since they can estimate plausible motion even in partially occluded shape parts, when combined with motion regularization such as ARAP~\cite{sorkine2007arap}.
Existing non-rigid reconstruction approaches build the deformation graph incrementally, i.e., frame-by-frame, which can lead to unstable graph configurations in the case of tracking errors.
In our approach, we predict a globally consistent deformation graph that can represent motion in all frames of the sequence, while being robust w.r.t.\ tracking errors present in challenging frames.

Sparse motion representations are common for human deformation modeling as well: the human skeletons used in \cite{loper2015smpl, cao2018openpose} are also instances of deformation graphs.
Some works have tried to extend human skeletons to more general objects, but with limited success.
In \cite{baran2007pinocchio}, a fixed generic skeleton is fitted to different object meshes, resulting in human-like re-animation of characters, but not general enough to be able to represent all degrees of freedom of general shapes.
Fixed hierarchical deformation graphs are used for differentiable non-rigid tracking in \cite{tretschk2019demea}, but a pre-computed graph template is required, with fixed connectivity on different coarseness levels. Thus, it is only applicable to specific object types (e.g., used for hand tracking).
Data-driven skeleton prediction has been introduced in \cite{xu2020rignet}, but it requires a dataset of manually designed skeletons as supervision, which is hard to obtain for general objects.
Our method, instead, estimates both deformation graph nodes and connectivity of general deformable objects in a self-supervised manner.

\paragraph{Implicit Surface Representation}
Representing surface geometry implicitly with a signed distance field (SDF) has been extensively used in the non-rigid reconstruction community.
An efficient algorithm for SDF grid construction from range images has been presented in \cite{curless1996volumetric} and extended to support non-rigid deformations in \cite{newcombe2015dynamicfusion}.
These methods rely on a discretized 3D grid to store the SDF, which can cause loss of detail, since grid resolution is limited by available memory.
A promising direction is to not use discretized grids at all, but instead represent the SDF function continuously using a multi-layer perceptron (MLP), as introduced in \cite{park2019deepsdf, sitzmann2019scene, chibane2020ifnets}. 
An implicit surface representation is used in \cite{huang2020arch} for accurate human reconstruction, where the SDF is estimated in a canonical T-pose space.
Since there exist no methods for estimation of canonical T-pose spaces for general non-rigid shapes, we instead base our method on the approach of Deng et al.~\cite{deng2019nasa}.
Assuming ground-truth dense body and skeleton tracking, they represent the human body with multiple MLPs, one for each bone and in its own canonical space, centered around the bone, therefore eliminating the need for a T-pose space.
In our general reconstruction approach, we estimate a deformation graph via self-supervision, and append an MLP to each deformation node to represent the surface of the observed object.

While most implicit reconstruction approaches do not produce consistent tracking, methods such as \cite{genova2019sif, genova2020ldif, tretschk2020patchnets} reconstruct objects in a patch-based manner and empirically observe consistency of patches across different deformations.
We compare our method, which leverages explicit consistency constraints, to these approaches to evaluate such implicit patch consistency.

\section{Method}

Given a sequence of signed distance fields observing a non-rigidly deforming surface, our method estimates the dense deformable motion in the sequence as well as reconstructs the geometry of the observed shape.
Specifically, we apply self-supervised learning on the sequence that we want to reconstruct.
A convolutional neural network that takes a signed distance field (SDF) as input is trained to predict a consistent deformation graph.
We call this neural network \emph{\OURS{}}, as it implicitly stores the deformation graphs of each frame.
Using the predicted graph node positions and orientations, we learn implicit functions to represent the shape of each graph node and, thus, the entire shape of the object.
The implicit functions are represented as a multi-layer perceptron (MLP). These MLPs take a 3D point centered around the node position as input and predict its signed distance value, defining the local geometry around the node.
Warping all node MLPs to every time step and interpolating their local part reconstruction results in an accurate \emph{implicit deformable shape reconstruction} without the need for an explicit canonical pose.
In addition to the sample point locations, the MLPs are conditioned on the predicted graph positions, which enables reconstruction of pose-dependent geometry detail.
Using Marching Cubes~\cite{marching_cubes}, the geometry can be extracted as a mesh at every time step, with dense correspondences estimated throughout the entire sequence.

\subsection{\OURS s}
A deformation graph consists of graph nodes and graph edges.
We represent the graph nodes $\mathcal{V}$ of each frame of the sequence implicitly by a neural network (\OURS{}).
The graph connectivity is explicitly stored for the entire sequence as an affinity matrix $\mathcal{E} \in \mathbb{R}^{N \times N}$ ($N = |\mathcal{V}|$).
A node $\mathpzc{v}\in \mathcal{V}$ is characterized by its 3D position  $\mathbf{v} \in \mathbb{R}^{3}$, rotation $R \in \mathbb{R}^{3 \times 3}$, importance weight $w \in \mathbb{R}$ (describing importance of the node, explained below), radius $r \in \mathbb{R}$ (describing the spatial influence of the node), and a local implicit shape function $f$.
We denote the set of node positions $\mathbf{V} = \{\mathbf{v}\}$, rotations $\mathbf{R} = \{R\}$, importance weights $\mathbf{W} = \{w\}$, radii $\mathbf{r} = \{r\}$, and shape functions $\mathbf{f} = \{f\}$, with $\mathcal{V} = (\mathbf{V}, \mathbf{R}, \mathbf{W}, \mathbf{r}, \mathbf{f})$.

To generate the signed distance fields needed for our method, we assume four calibrated cameras. The depth maps at each time step $k$ are back-projected into a common coordinate system and converted into a signed distance field $\mathbf{S}_k$ of dimension $64^3$ using static volumetric reconstruction \cite{kazhdan2013screenedpoisson}. Note that due to occlusions this representation is partial, thus only an approximate signed distance field is used for our deformation graph prediction.
Based on this input, we estimate $(\mathbf{V}_k, \mathbf{R}_k, \mathbf{W}_k)$ using a \OURS{} ($\mathrm{NDG}$) which is based on a 3D convolutional neural network (see supplemental material for architecture details):
\begin{equation*}
(\mathbf{V}_k, \mathbf{R}_k, \mathbf{W}_k) = \mathrm{NDG} (\mathbf{S}_k).
\end{equation*}
The radii $\mathbf{r}$ of the graph nodes as well as the graph node affinities $\mathcal{E}$ are jointly optimized over the entire input sequence.
In addition to the affinity matrix, we also store the average edge lengths (node-to-node distances) $\mathbf{D} \in \mathbb{R}^{N \times N}$, which are used for regularization.
For every graph node, we also optimize for a local MLP which is used to represent the surface of the object (see Sec~\ref{subsec:implicit_representation}).

We define a fixed number of graph nodes ($N=100$) in our experiments; note that this is an upper bound on the effective number of nodes, since the importance weights allow eliminating the effect of redundant nodes, making our method applicable to shapes of different size and structure complexity.
To achieve a consistent graph node prediction via self-supervised training, we employ the following constraints for each time-step $k$.

\paragraph{Graph coverage loss.}
A deformation graph should cover the entire object to ensure that every deformable part can be represented while simultaneously enforcing that free space is not covered.
To this end, we employ a loss that encourages the coverage of the shape by the node centers (w.r.t.\ their radii).
We define the influence of a node (with position $\mathbf{v}$, radius $r > 0$, and importance weight $w > 0$) on a point $\mathbf{x} \in \mathbb{R}^3$ using a \emph{weighted Gaussian function}:
\begin{equation*}
\mathcal{G} (\mathbf{x}, \mathbf{v}, r, w) = w \cdot \exp{\left(\frac{||\mathbf{x} - \mathbf{v}||^2_2}{r^2}\right)}
\end{equation*}
The coverage of a point $\mathbf{x} \in \mathbb{R}^3$ is computed by summing the corresponding contributions of all nodes, and applying a \emph{sigmoid} to encourage a fast transition from covered (where coverage value is $1$) to free space (where coverage value should be $0$), enabling more accurate surface coverage:
\begin{equation*}
\mathcal{C} (\mathbf{x}, \mathbf{V}_k, \mathbf{r}, \mathbf{W}_k) = \sigma \left(s \left( \left(\sum_{\mathbf{v}, r, w} \mathcal{G} (\mathbf{x}, \mathbf{v}, r, w) \right) - d\right) \right)
\end{equation*}
We empirically set $d = 0.07$ and $s = 100.0$. 
To compute the coverage loss, we sample points $P_{\mathrm{un}}$ uniformly in the shape's bounding box and points $P_{\mathrm{ns}}$ near the surface region. Points are assigned coverage value of $c = 0$ if they are visible in at least one of the cameras, otherwise they are assigned $c = 1$. 
The coverage loss then compares predicted coverage of these point samples with the pre-computed coverage using an $\ell_2$ loss:
\begin{align*}
\mathcal{L}_{\mathrm{coverage}} =& \lambda_{\mathrm{un}} \sum_{(\mathbf{x}, c) \in P_{\mathrm{un}}} || \mathcal{C} (\mathbf{x}, \mathbf{V}_k, \mathbf{r}, \mathbf{W}_k) - c ||^2_2\; + \\
&\lambda_{\mathrm{ns}} \sum_{(\mathbf{x}, c) \in P_{\mathrm{ns}}} || \mathcal{C} (\mathbf{x}, \mathbf{V}_k, \mathbf{r}, \mathbf{W}_k) - c ||^2_2
\end{align*}

\paragraph{Node interior loss.}
In addition to the graph coverage loss, we require the node positions to be predicted inside the shape.
If any node's position $\mathbf{v}$ is predicted outside the observed surface $\mathbf{S}_k$, i.e., in the SDF region with positive signed distance value, we penalize it to encourage the node's position to be inside the surface:
\begin{equation*}
\mathcal{L}_{\mathrm{interior}} = \sum_{\mathbf{v} \in \mathbf{V}_k} \max \left( \mathrm{interp} (\mathbf{S}_k, \mathbf{v}), 0 \right)
\end{equation*}
Here $\mathrm{interp}(\mathbf{S}_k, \mathbf{v})$ is the trilinear interpolation of $\mathbf{S}_k$ at $\mathbf{v}$.

\paragraph{Affinity consistency loss.}
We also optimize for a global affinity matrix $\mathcal{E} = \{e_{ij}\; |\; i\in[1,N], j\in[1,N]\}$ representing node-to-node affinities across the entire input sequence.
We compute node-to-node Euclidean distances $||\mathbf{v}_i^k - \mathbf{v}_j^k||_2$ at each frame $k$, and weight them by connectivity weights $e_{ij}$; this should remain consistent over the whole sequence (relative loss, preserving edge length) and relatively small (absolute loss, preferring close-by connections).
To ensure global distance consistency, we additionally optimize over average node-to-node distances $d_{ij}$, resulting in the affinity loss:
\begin{align*}
\mathcal{L}_{\mathrm{affinity}} =& \lambda_{\mathrm{rel}} \sum_{i \ne j} e_{ij} \Big| d_{ij}^2 - ||\mathbf{v}_i^k - \mathbf{v}_j^k||^2_2 \Big|_1 + \\
&\lambda_{\mathrm{abs}} \sum_{i \ne j} e_{ij} ||\mathbf{v}_i^k - \mathbf{v}_j^k||^2_2
\end{align*}

\paragraph{Sparsity loss.}
We enforce a sparse connectivity of the graph.
Specifically, each node can have up to $K$ neighbors ($K = 2$ in our setting); we use a (soft) loss to encourage these neighbors to be different.
To achieve this, we optimize over a set of matrices $\mathbf{A}_1, \dots, \mathbf{A}_K \in \mathbb{R}^{N \times N}$, and construct $\mathcal{E} \in \mathbb{R}^{N \times N}$ as:
\begin{equation*}
\mathcal{E} = \frac{1}{K} \sum_{i = 1}^{K} \mathrm{softmax} (\mathbf{A}_i)
\end{equation*}
We use $\mathrm{softmax}$ over the rows of matrix $\mathbf{A}_i$ to guarantee all affinity elements of a node to be positive and add up to $1$.
To enforce unique graph neighbors, a sparsity loss is employed, encouraging different matrices $\mathbf{A}_i$ to produce different neighbors: %
\begin{equation*}
\mathcal{L}_{\mathrm{sparsity}} = \sum_{l \ne m} ||\mathrm{softmax} (\mathbf{A}_l) \odot \mathrm{softmax} (\mathbf{A}_m)||^2_F
\end{equation*}
We use $\odot$ to denote the element-wise product. %

\subsection{Global Deformation Optimization}

We compute deformation between any pair of frames by interpolating the nodes' relative motions (translations and rotations), weighted by their influences $\mathcal{G}$.
For a source frame $s$ and target frame $t$, the warping of point $\mathbf{x} \in \mathbb{R}^3$ from frame $s$ to frame $t$ is defined as:
\begin{equation*}
\mathcal{W}_{s \to t} (\mathbf{x}) = \sum_{i = 1}^N \mathcal{G} (\mathbf{x}, \mathbf{v}_i^s, r_i, w_i^s) (\mathbf{R}_i^t
(\mathbf{R}_i^s)^\mathrm{T}  (\mathbf{x} - \mathbf{v}_i^s) + \mathbf{v}_i^t)
\end{equation*}
We denoted parameters at the source frame with $(\cdot)^s$ and at the target frame with $(\cdot)^t$.
We use the Embedded Deformation formulation \cite{sumner2007ed} to parameterize frame-to-frame deformation, but instead of fixed-radius skinning we employ node influence $\mathcal{G}$ as the skinning weight, which enables different skinning effects for every node as well as frame-adaptive skinning, i.e., skinning can change depending on the deformation.
To ensure globally consistent deformation, we employ a per-frame viewpoint consistency loss and a surface consistency loss.

\paragraph{Viewpoint consistency loss.}
Since input observations may see very different views, we enforce a viewpoint consistency loss for consistent graph node predictions across varying views.
To this end, for each frame $k$, the rotated 3D input $\mathbf{S}_k$ should produce consistent graph node positions $\mathbf{V}_k$, rotations $\mathbf{R}_k$ and importance weights $\mathbf{W}_k$.
In our experiments, we only consider view rotations around the y-axis, since the camera setup is arranged in the x-z plane.
In each batch, we sample two random angles $\alpha$ and $\beta$ for every sample, and compute rotated inputs $\pi_\alpha (\mathbf{S}_k)$ and $\pi_\beta (\mathbf{S}_k)$ by trilinear re-sampling of input SDF grid $\mathbf{S}_k$ using rotated grid indices.
Viewpoint consistency is then measured by:
\begin{equation*}
\mathcal{L}_{\mathrm{vc}} = ||\pi^{-1}_\alpha \mathrm{NDG} (\pi_\alpha (\mathbf{S}_k)) - \pi^{-1}_\beta \mathrm{NDG} (\pi_\beta (\mathbf{S}_k))||^2_2
\end{equation*}
where the function $\pi^{-1}_\phi$ corrects for the input rotation of angle $\phi$: $\pi^{-1}_\phi (\mathbf{V}_k, \mathbf{R}_k, \mathbf{W}_k) = (R_\phi^{\mathrm{T}} \mathbf{V}_k, R_\phi^{\mathrm{T}} \mathbf{R}_k, \mathbf{W}_k)$.

\paragraph{Surface consistency loss.}
Surface points from a source frame $s$ should, after deformation to a target frame $t$, align well with the target frame's SDF grid $\mathbf{S}_t$.
We sample surface points $P_{\mathrm{s}}$ in the source frame and warp them to the target frame using the predicted deformation, to trilinearly interpolate the target grid $\mathbf{S}_t$, encouraging surface points to be warped to near zero (surface) SDF values:
\begin{equation*}
\mathcal{L}_{\mathrm{sc}} = \sum_{\mathbf{x} \in P_{\mathrm{s}}} (\mathrm{interp} (\mathbf{S}_t, \mathcal{W}_{s \to t} (\mathbf{x})))^2 
\end{equation*}
This consistency loss is computed between pairs of samples in the batch, with uniformly sampled batch samples.

\subsection{Implicit Surface Reconstruction}\label{subsec:implicit_representation}
We represent the surface of the object as an implicit function.
Specifically, each graph node $i$ defines local geometry over the influence of that node, with an implicit function $f_i$, represented by an MLP.
This MLP takes a location in the local space as input and outputs an SDF value.
Any point $\mathbf{x} \in \mathbb{R}^3$ in the current frame $k$ can be transformed to the local coordinate system of node $i$ as
$\mathcal{W}_{k,i}^{-1} (\mathbf{x}) = \mathbf{P}_{\mathrm{enc}} ( (\mathbf{R}_i^k)^\mathrm{T}  (\mathbf{x} - \mathbf{v}_i^k) )$.
$\mathbf{P}_{\mathrm{enc}}: \mathbb{R}^3 \to \mathbb{R}^F$ denotes positional encoding that transforms 3D local coordinates to a high-dimensional frequency domain (in our case $F = 30$), as presented in \cite{mildenhall2020nerf}.
Inspired by Deng et al.~\cite{deng2019nasa}, we condition each $f_i$ on the predicted input frame's graph parameters, such that they can encode pose-specific geometry details.
We train a linear layer $\Pi_i(\cdot)$ to select a sparse pose code (of dimension $D=32$) for every $f_i$ from the graph predictions $\mathrm{NDG} (\mathbf{S}_k)$. %
Given this input of dimension $D + F$, we use $8$ linear layers with feature dimension of $32$, a leaky ReLU (with negative slope of $0.01$) as activation function, and skip connections between the input and the 6th linear layer.

We compute the full surface reconstruction $\mathcal{S}_k$ as an SDF created from interpolating the SDF output values of each local MLP $f_i$, using the aforementioned skinning weights and transformations to the current frame by the estimated nodes' rotations and translations:
\begin{equation*}
\mathcal{S}_k (\mathbf{x}) = \sum_{i = 1}^N \mathcal{G} (\mathbf{x}, \mathbf{v}_i^k, r_i, w_i^k) f_i( \mathcal{W}_{k,i}^{-1} (\mathbf{x}), \Pi_i (\mathrm{NDG} (\mathbf{S}_k)) )
\end{equation*}
This operation is efficiently implement using group convolutions.
During training, we use the same point samples $P_{\mathrm{un}}$ and $P_{\mathrm{ns}}$ as for the graph coverage loss, sampled uniformly and near the surface, but instead of the $0/1$ coverage values we use their approximate SDF values.
We then optimize for $\{f_i\}$ using the SDF reconstruction loss:
\begin{equation*}
\mathcal{L}_{\mathrm{recon}} = \sum_{(\mathbf{x}, \mathrm{sdf}) \in P_{\mathrm{un}} \cup P_{\mathrm{ns}}} | \mathcal{S}_k (\mathbf{x}) - \mathrm{sdf} |_1
\end{equation*}

\subsection{Training Details}\label{subsec:training_details}
We use the Adam solver~\cite{adam} with momentum of $0.9$ to optimize the complete loss:
\begin{align*}
\mathcal{L} &= \mathcal{L}_{\mathrm{coverage}} + \lambda_{\mathrm{interior}} \mathcal{L}_{\mathrm{interior}} + \mathcal{L}_{\mathrm{affinity}} + \\
&\lambda_{\mathrm{sparsity}} \mathcal{L}_{\mathrm{sparsity}} + \lambda_{\mathrm{vc}} \mathcal{L}_{\mathrm{vc}} + \lambda_{\mathrm{sc}} \mathcal{L}_{\mathrm{sc}} + \lambda_{\mathrm{recon}} \mathcal{L}_{\mathrm{recon}}
\end{align*}
Our method is trained in two stages. We initially train the CNN encoder with all losses except the reconstruction loss, and afterwards train the multi-MLP network using only the reconstruction loss, with the CNN encoder frozen.

The CNN encoder is trained for $500$k iterations with a learning rate of $5\mathrm{e}^{-5}$ and a batch size of $16$; we balance the losses with $\lambda_{\mathrm{un}} = 1.0$, $\lambda_{\mathrm{ns}} = 0.1$, $\lambda_{\mathrm{interior}} = 1.0$, $\lambda_{\mathrm{rel}} = 0.1$, $\lambda_{\mathrm{abs}} = 0.1$, $\lambda_{\mathrm{sparsity}} = 1\mathrm{e}^{-8}$, $\lambda_{\mathrm{vc}} = (10.0, 1.0, 1\mathrm{e}^{-4})$ (for graph node's position, weight and rotation, respectively) and $\lambda_{\mathrm{sc}} = 1\mathrm{e}^{-6}$.
Every 50k iterations we increase the loss weights $\lambda_{\mathrm{rel}}$, $\lambda_{\mathrm{abs}}$, $\lambda_{\mathrm{sparsity}}$, $\lambda_{\mathrm{sc}}$ by a factor of $10$, up to maximum weights $\lambda_{\mathrm{rel}}^{\mathrm{max}} = 10000.0$, $\lambda_{\mathrm{abs}}^{\mathrm{max}} = 1.0$, $\lambda_{\mathrm{sparsity}}^{\mathrm{max}} = 1\mathrm{e}^{-3}$ and $\lambda_{\mathrm{sc}}^{\mathrm{max}} = 1000.0$.

The multi-MLP network is trained for $500$k iterations  with a learning rate of $5\mathrm{e}^{-4}$ and a batch size of $4$, only based on the reconstruction loss with $\lambda_{\mathrm{recon}} = 1.0$.

\section{Results}

\begin{figure*}
\includegraphics[width=\linewidth]{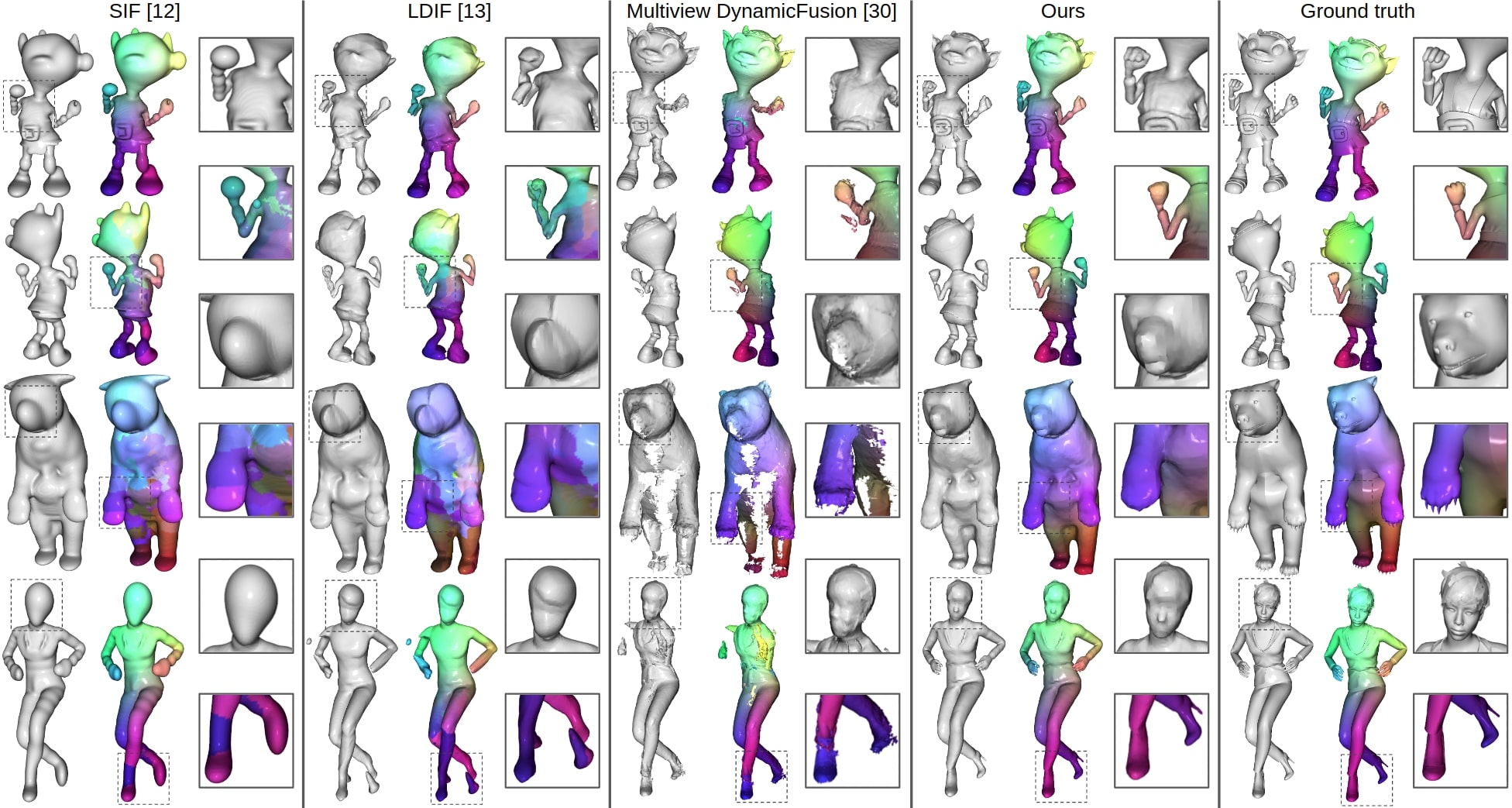}%
\caption{
We qualitatively compare our method to the baseline methods on synthetic data.
Colors represent corresponding locations in the first frame of the sequence and, thus, giving a notion of the tracking quality and consistency.
Our approach outperforms state of the art in both reconstruction and deformation tracking quality.
}
\label{fig:qualitative-comparison-synthetic}
\end{figure*}

\begin{table}[]
\centering
\begin{tabular}{lccccc}
\toprule
\textbf{Method} & \textbf{Chamfer} & \textbf{EPE3D} \\ 
\midrule
SIF \cite{genova2019sif} & $1.12$ & $8.56$ \\
LDIF \cite{genova2020ldif} & $2.41$ & $10.40$  \\
Multiview DynamicFusion \cite{newcombe2015dynamicfusion} & $2.19$ & $3.06$  \\ 
\midrule
Ours: GRAPH & $0.50$ & $8.93$ \\
Ours: GRAPH + AO & $0.46$ & $8.03$ \\
Ours: GRAPH + AO + VC & $0.44$ & $4.12$ \\
\midrule
Ours: GRAPH + AO + VC + SC & $\mathbf{0.40}$ & $\mathbf{1.16}$ \\ 
\bottomrule
\end{tabular}
\caption{\label{tab:quantitative-comparison} We show quantitative comparisons with state-of-the-art approaches, evaluating geometry using Chamfer distance ($\times 10^{-4}$), and deformation using EPE3D ($\times 10^{-2}$). We also include an ablation study of different components of our method: GRAPH~=~coverage and interior losses, AO~=~affinity optimization with affinity consistency and sparsity, VC~=~viewpoint consistency, SC~=~surface consistency.}
\end{table}

\begin{figure*}
\includegraphics[width=\linewidth]{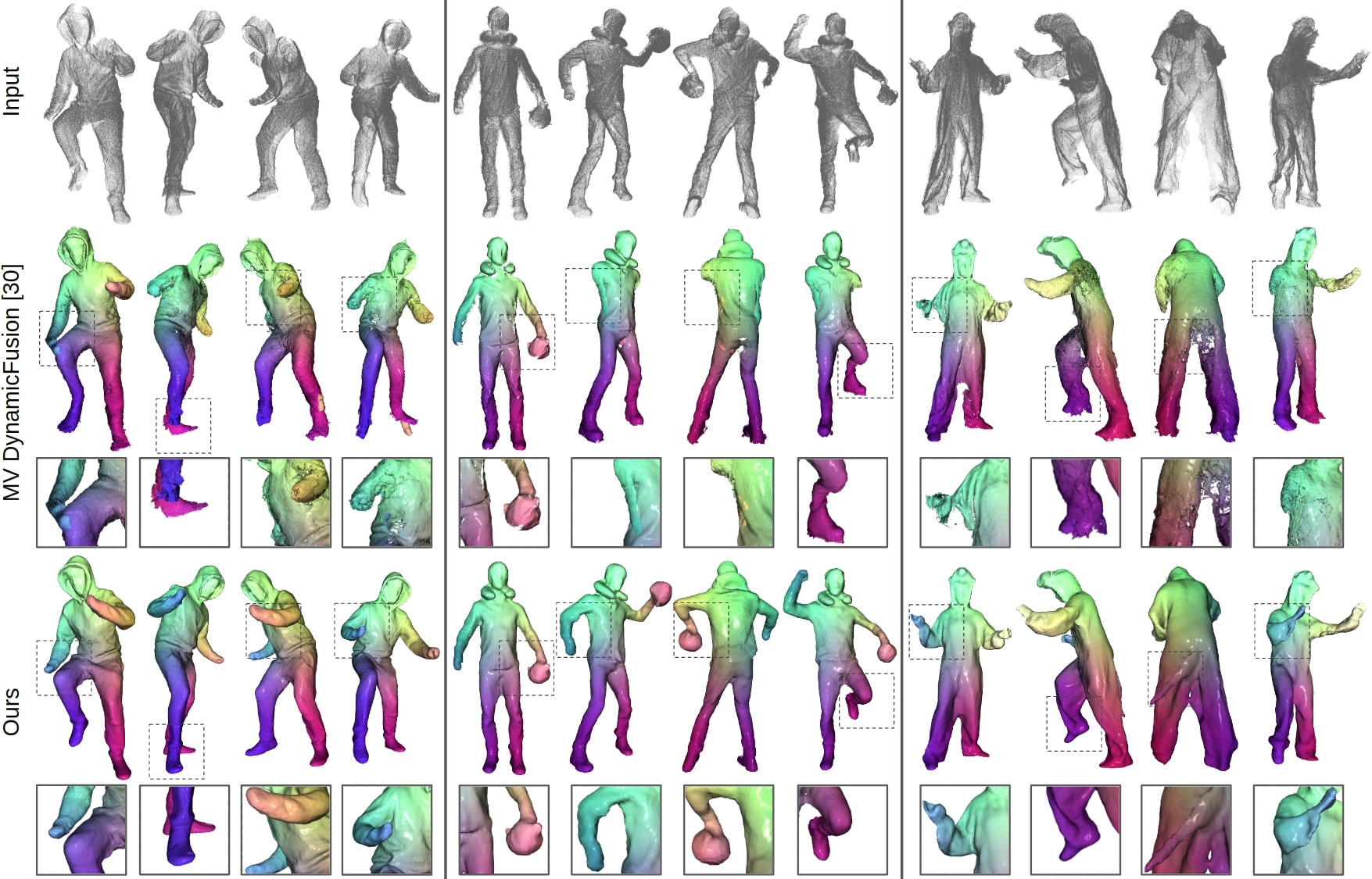}%
\caption{
We show qualitative comparisons of our method with multi-view DynamicFusion~\cite{wang2018dynamic} on real sequences captured by four Kinect Azure sensors. Colors represent corresponding locations in the first frame of the sequence to visualize the tracking quality and consistency.
}
\label{fig:qualitative-comparison-real}
\end{figure*}

To evaluate our proposed approach, we conducted a series of experiments on real and synthetic recordings where ground truth data is available.

\paragraph{Evaluation on Synthetic Data}
In order to quantitatively and qualitatively evaluate our method, we make use of synthetic human-like and character sequences from the DeformingThings4D dataset~\cite{yang20204dcomplete}.
To mimic our real data capture setup, we render 4 fixed depth views for every frame of synthetic animation, and generate SDF grids from these 4 views.  
Quantitative evaluation is executed on three sequences, including human, character and bear motion, as shown in Fig.~\ref{fig:qualitative-comparison-synthetic}.

We compare our approach to state-of-the-art network-based reconstruction methods, SIF~\cite{genova2019sif} and LDIF~\cite{genova2020ldif}, as well as to the non-rigid reconstruction approach DynamicFusion~\cite{newcombe2015dynamicfusion}, which we extend to the multi-view domain.
Both SIF and LDIF execute part-based shape reconstruction, predicting an ellipsoid for every shape part.
While the methods do not explicitly focus on deformation tracking, we add dense tracking computation via ellipsoid motion interpolation between two frames, represented by relative translation and rotation.
Our multi-view DynamicFusion baseline is a specialized framework for both non-rigid tracking and reconstruction, with coarse-to-fine multi-frame alignment using depth iterative closest point (ICP) correspondences and non-rigid volumetric fusion.

In Tab.~\ref{tab:quantitative-comparison}, we evaluate the deformation estimation and geometry reconstruction performance of these methods in comparison to our approach.
The depth data of every sequence is normalized such that the largest bounding box side length is equal to $1.0$.
All numbers are listed w.r.t.~this unit cube, thus, being independent to the scale of the objects. %
The geometry reconstruction is evaluated using \emph{L2 Chamfer distance}, which computes average squared bi-directional point-to-point distance between reconstructed and ground truth geometry for every time step, thus evaluating accuracy and completeness of geometry.
For deformation evaluation, we uniformly sample $10$ keyframes per sequence (with fixed step size), and evaluate dense deformation from each of these keyframes to any other frame by deforming the ground truth points at a \textbf{}keyframe to every other frame.
As deformation metric we use the \emph{End-Point-Error (EPE3D)}, measuring average L2 distance between estimated keyframe-to-frame deformations and ground truth deformations.

Among the baselines the best reconstruction performance (lower Chamfer distance) is achieved by SIF~\cite{genova2019sif}, while multi-view DynamicFusion~\cite{newcombe2015dynamicfusion} obtains better deformation tracking performance (lower EPE3D).
However, our approach outperforms all methods on both reconstruction and deformation tracking metrics, achieving $64$\% better reconstruction and $62$\% better deformation tracking results.
The improvement is also clearly visible in the qualitative comparisons shown in Fig.~\ref{fig:qualitative-comparison-synthetic}.
We visualize tracking performance by a color gradient that is defined in the initial frame of each sequence.
Specifically, each point is given a color value w.r.t.~its location in the bounding box of the first frame.
With perfect tracking and reconstruction, a specific point on the surface will have the same color throughout the sequence. Errors in tracking result in wrong surface colors.
The methods SIF~\cite{genova2019sif} and LDIF~\cite{genova2020ldif} produce less accurate geometry reconstruction with tracking outliers under larger deformation (e.g., flipped legs in the human sequence).
Multi-view DynamicFusion~\cite{newcombe2015dynamicfusion} suffers from incomplete geometry because of the incremental graph construction and surface integration. DynamicFusion can also not recover from tracking failures, i.e., shape parts where frame-to-frame tracking failed start to disappear and are integrated at a different location.
In contrast, our method is robust in the case of larger deformations and produces complete and accurate geometry reconstruction.

\paragraph{Evaluation on Real Data.}
Our real data capture setup consists of $4$ Kinect Azure sensors with hardware synchronization.
The cameras are calibrated with a checkerboard using OpenCV~\cite{lepetit2009epnp} and an additional refinement procedure based on ICP~\cite{medioni1992}.
Before recording an actual sequence, we record the background to compute the floor plane using PCA.
During capture, we filter out floor points and background points, i.e., all points outside of a cylinder with diameter $1.8$~m and height $2.5$~m.
We use the wide-field-of-view depth capture setting with a resolution of $1024\times1024$ pixels, at the highest available frame-rate of $15$ FPS for this resolution.
In Fig.~\ref{fig:qualitative-comparison-real}, we show a comparison between the multi-view DynamicFusion approach~\cite{newcombe2015dynamicfusion} and ours.
Our approach achieves considerably more accurate deformation tracking (color is retrieved from the first frame) while also producing more complete and accurate geometry reconstruction.
More qualitative results are shown in the accompanying video.

\paragraph{Ablation study of network components.}
To evaluate specific parts of our method, we employ an ablation study.
Specifically, we analyzed the performance of our method by performing optimization without using certain losses: without affinity related losses (affinity consistency and sparsity losses), viewpoint consistency loss and surface consistency loss.
As shown in Tab.~\ref{tab:quantitative-comparison}, using these additional losses vastly improves the method's performance.
Especially, it results in a much lower EPE3D error, and, thus, in globally consistent tracking performance.

\paragraph{Limitations.}
Using our globally consistent \OURS{}, we show state-of-the-art tracking and reconstruction quality.
Currently, our quality is limited by the input, i.e., a $64^3$ SDF grid.
Sparse 3D convolutions~\cite{3DSemanticSegmentationWithSubmanifoldSparseConvNet} could be applied to cope with higher resolutions.
Our approach focuses on the tracking and reconstruction of the geometry, and not the texture.
A texture on top of the tracked geometry could be estimated (similar to the color scheme that we show in the results figures) and additional losses based on this texture could be employed.
Based on our approach, we believe that there is a potential of several follow-up works with adapted loss formulations and additional features like color reconstruction.

\section{Conclusion}
We introduced \OURS{} which allows to reconstruct and track non-rigidly deforming objects in a globally consistent fashion.
It is enabled by a neural network that implicitly stores the deformation graph of the object.
The network is trained with losses on global consistency, resulting in tracking and reconstruction quality that surpasses the state of the art by more than $60$\% w.r.t.\ the respective metrics.
We believe that our global optimization of non-rigid motion will be a stepping stone to learn data-driven priors in the future.

\section*{Acknowledgments}

This work was supported by the ZD.B (Zentrum Digitalisierung.Bayern), a TUM-IAS Rudolf M\"o{\ss}bauer Fellowship, the ERC Starting Grant Scan2CAD (804724), and the German Research Foundation (DFG) Grant Making Machine Learning on Static and Dynamic 3D Data Practical.

\begin{appendix}
\section*{Appendix}

In this appendix, we provide further details about our proposed neural deformation graphs.
Specifically, we describe the network architectures in detail (Sec.~\ref{sec:network_architecture}), give additional training information (Sec.~\ref{sec:training_details}), and present our capture setup used for recording real non-rigid motion sequences using Kinect Azure sensors (Sec.~\ref{sec:real_data_capture_setup}).

\section{Network Architecture}
\label{sec:network_architecture}
\begin{figure}
    \centering
    \includegraphics[width=\linewidth]{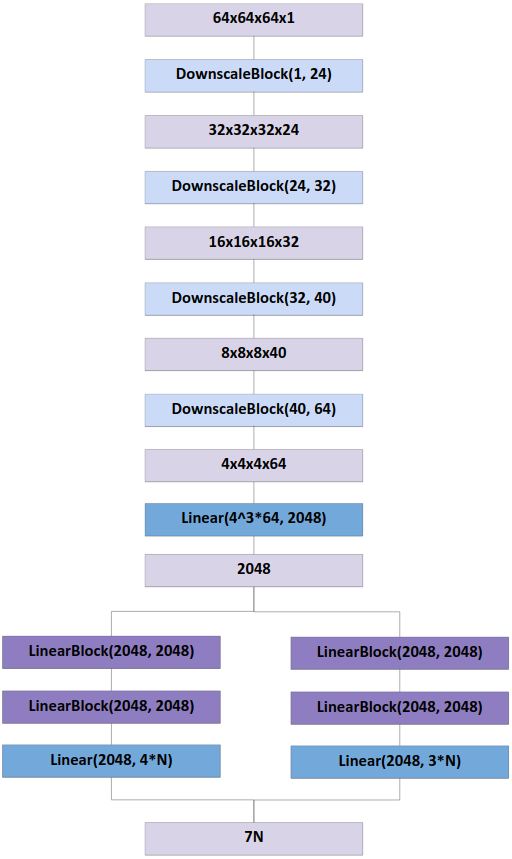}
    \caption{
    Our neural deformation graph is represented as a 3D CNN which takes as input a $64^3$ SDF grid and outputs $7N$ graph node parameters.
    The architectures of the single computation blocks are detailed in Fig.~\ref{fig:downscale_block} and Fig.~\ref{fig:linear_block}.
    }
    \label{fig:3d_cnn_encoder}
\end{figure}

\begin{figure}
    \centering
    \includegraphics[width=0.55\linewidth]{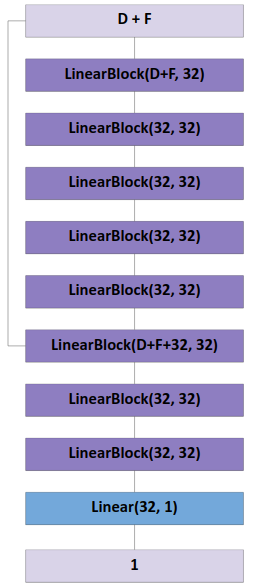}
    \caption{
    We represent the surface of an object using an implicit function (SDF) which is based on multi-layer perceptrons (MLPs).
    Each graph node MLP takes as input a $(D+F)$-dimensional vector (consisting of the pose code and the sample position, represented with positional encoding \cite{mildenhall2020nerf} in local coordinates) and outputs the SDF value.
    }
    \label{fig:mlp}
\end{figure}

Our method is composed of two learned components; the neural deformation graph  based on a 3D CNN (shown in Fig.~\ref{fig:3d_cnn_encoder}) and a set of multi-layer perceptrons (MLPs) which are used to implicitly represent the surface (see Fig.~\ref{fig:mlp}).

\paragraph{Neural Deformation Graph.}

\begin{figure}
    \centering
    \includegraphics[width=0.8\linewidth]{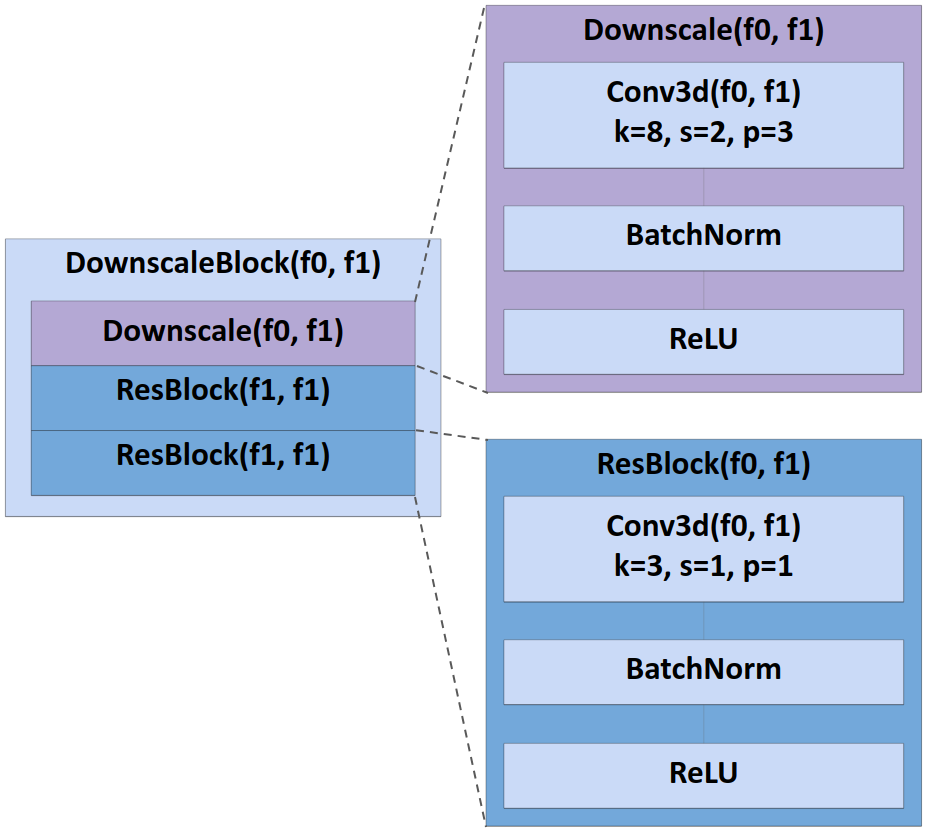}
    \caption{
    The downscale block used in our 3D CNN architecture (see Fig.~\ref{fig:3d_cnn_encoder}) reduces the spatial dimension by $2$.
    }
    \label{fig:downscale_block}
\end{figure}

The neural deformation graph implicitly stores the deformation graphs for each input frame $k$.
The underlying 3D CNN encoder takes a dense $64^3$ SDF grid $\mathbf{S}_k$ as input and outputs the $(3 + 3 + 1)N$ dimensional vector, representing position, rotation (in axis-angle format) and importance weight for each graph node ($N$ being the number of graph nodes).
The input grid is encoded to a spatial dimension of $4^3$ and a feature dimension of $64$ using $4$ blocks of convolutional downsampling with stride of $2$, followed by $2$ residual units (with ReLU activation unit and batch norm).
The downscale operation is detailed in Fig.~\ref{fig:downscale_block}.
We use a linear layer to convert the $4^3 \cdot 64$ features to dimension $2048$.
This feature vector is input to two independent heads, each composed of $3$ linear layers with feature dimension of $2048$ and a leaky ReLU as activation unit with negative slope of $0.01$ (no activation function after the last linear layer).
One head predicts graph node rotations $\mathbf{R}_k$ of dimension $3N$, while the other head predicts graph node positions $\mathbf{V}_k$ and weights $\mathbf{W}_k$ (total dimension of $(3+1)N$).
We concatenate both predictions, which results in a graph embedding of dimension $7N$.
For all our experiments, we use $N = 100$.

\begin{figure}
    \centering
    \includegraphics[width=0.4\linewidth]{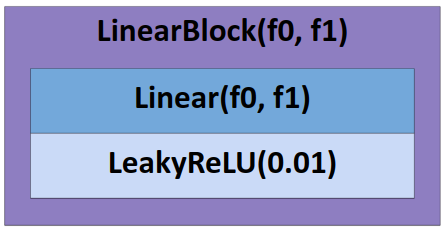}
    \caption{
    Each linear block of the 3D CNN architecture (see Fig.~\ref{fig:3d_cnn_encoder}) applies a linear layer, followed by a leaky ReLU with negative slope of $0.01$.
    }
    \label{fig:linear_block}
\end{figure}

\begin{figure}
    \centering
    \includegraphics[width=0.45\linewidth]{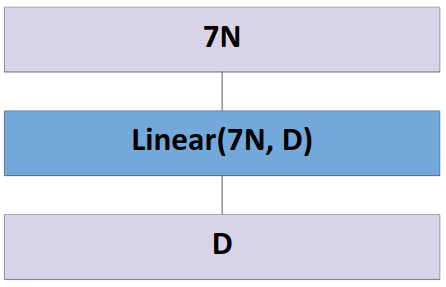}
    \caption{
    To compute the sparse pose code of the graph, we use a single linear projection layer that converts the graph parameters of dimension $7N$ to a code of dimension $D$ (we use $N = 100$ and $D = 32$).
    This pose code is input to the local MLP that represents the pose-dependent local surface.
    }
    \label{fig:projection_layer}
\end{figure}

\paragraph{Implicit Surface Representation}
For each graph node $i$, we train a separate MLP $f_i$ that predicts the signed distance field around the node, as presented in Fig.~\ref{fig:mlp}.
As explained in the main paper, the complete shape geometry is reconstructed by warping each node's local SDF using estimated node deformations and interpolating the local SDFs using the estimated node interpolation weights (computed from node radii and importance weights).
Each MLP takes a 3D position as input, and outputs an SDF value for the given position.
The position is represented with positional encoding~\cite{mildenhall2020nerf} of dimension $F = 30$.
To reconstruct pose-dependent details, the MLP is conditioned on a pose code. This $D$-dimensional pose code ($D = 32$ in our experiments) is computed by inputting the current deformation graph predictions through a linear projection layer $\Pi_i$ shown in Fig.~\ref{fig:projection_layer}.
Thus, in total, the MLP takes an input of dimension $D + F$, and using $8$ linear layers with feature dimension of $32$, it outputs the SDF value of the corresponding node. 
There is a skip connection between the input and the sixth linear layer, and a leaky ReLU with negative slope of $0.01$ is applied after each linear layer, as shown in Fig.~\ref{fig:linear_block}.

\begin{figure}
    \centering
    \includegraphics[width=\linewidth]{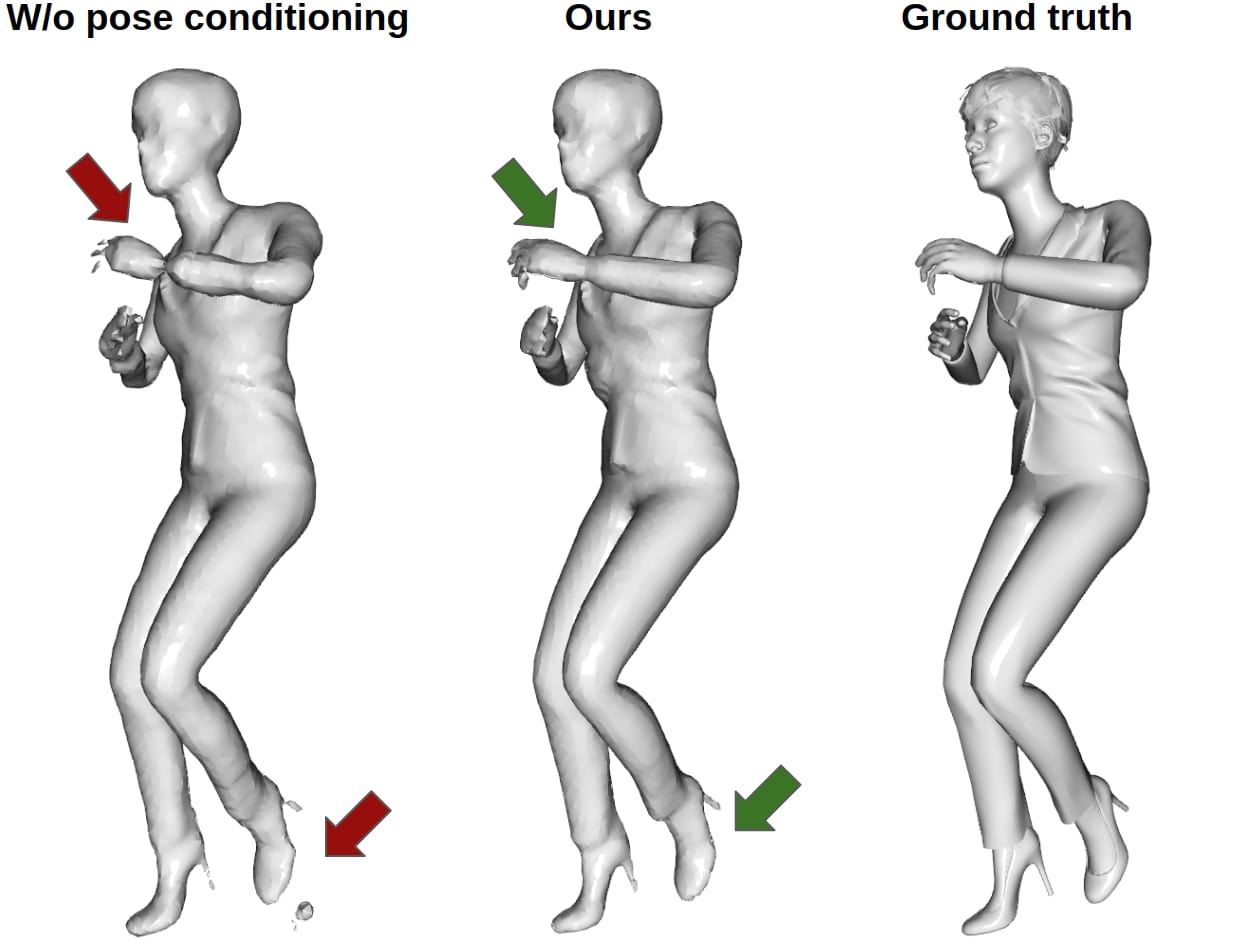}
    \caption{
    We compare our implicit reconstruction with pose conditioning (middle) to without pose conditioning (left).
    Pose conditioning clearly improves reconstruction performance in regions of very strong deformation.
    }
    \label{fig:pose_conditioning_comparison}
\end{figure}

In Tab.~\ref{tab:pose_conditioning}, we evaluate the influence of the pose codes with respect to the reconstruction quality measured using a Chamfer distance.
The pose conditioning improves the Chamfer distance by a large margin, which is also visible in the qualitative comparison in Fig.~\ref{fig:pose_conditioning_comparison}.

\begin{table}[]
    \centering
    \begin{tabular}{lc}
        \toprule
        \textbf{Approach} & \textbf{Chamfer} \\ 
        \midrule
        Without pose conditioning & $0.46$ \\
        With pose conditioning & $0.40$ \\
        \bottomrule
    \end{tabular}
    \caption{Evaluation of the pose conditioning on reconstructions on synthetic data using Chamfer distance ($\times 10^{-4}$).}
    \label{tab:pose_conditioning}
\end{table}

\begin{figure}
    \centering
    \includegraphics[width=\linewidth]{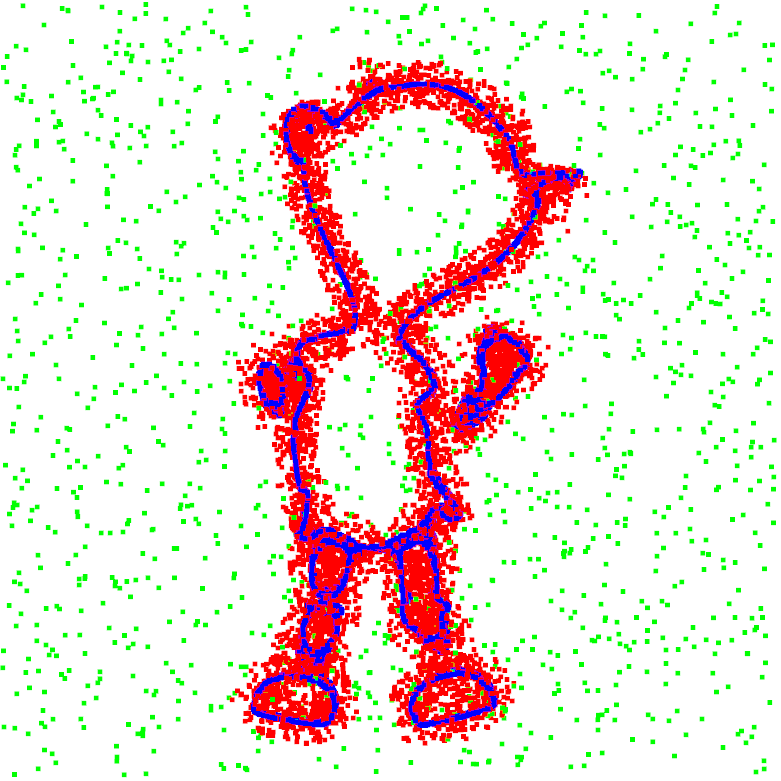}
    \caption{
    We visualize the uniform samples $P_{\mathrm{un}}$ (green), near surface samples $P_{\mathrm{ns}}$ (red) and the surface samples $P_{\mathrm{s}}$ (blue) for a slice of samples with $z \in [-0.01, 0.01]$ at one frame of the character sequence shown in the Fig.~\ref{fig:teaser} of the main paper.
    }
    \label{fig:sampling_visualization}
\end{figure}

\section{Training Details}
\label{sec:training_details}
Our method's training process is described in Sec.~\ref{subsec:training_details} of the main paper.
In this section, we specify the sampling strategy during training of both the neural deformation graph and the implicit surface representations (see visualization of samples in Fig.~\ref{fig:sampling_visualization}).

When training the neural deformation graph, we use $|P_{\mathrm{un}}| = 3000$ uniform point samples, $|P_{\mathrm{ns}}| = 3000$ near surface point samples and $|P_{\mathrm{s}}| = 3000$ surface point samples, sampled randomly for each batch from $100$k pre-processed point samples.
For the graph coverage loss, we apply an additional weight of $10.0$ for interior point samples (determined by the SDF sign).

To train our multi-MLP network that implicitly represents the surface, we use $|P_{\mathrm{un}}| = 1500$ uniform point samples and $|P_{\mathrm{ns}}| = 1500$ near surface point samples.
Note that this reduced set of samples is applied to satisfy memory limits of our used GPU (Nvidia Geforce 2080Ti).
We use a truncation of $0.1$ (in normalized units of the object in the unit cube) for the signed distance field used for the reconstruction loss.

\newpage
Note, when interpolating the SDF grid for the node interior loss, in the case that a graph node's position is predicted outside the grid, we define the out-of-grid loss by an $\ell_2$-distance to the nearest bounding box corner. 
This encourages graph node positions to be always predicted inside the shape's bounding box.

\section{Real Data Capture Setup}
\label{sec:real_data_capture_setup}

\begin{figure}
    \centering
    \includegraphics[width=\linewidth]{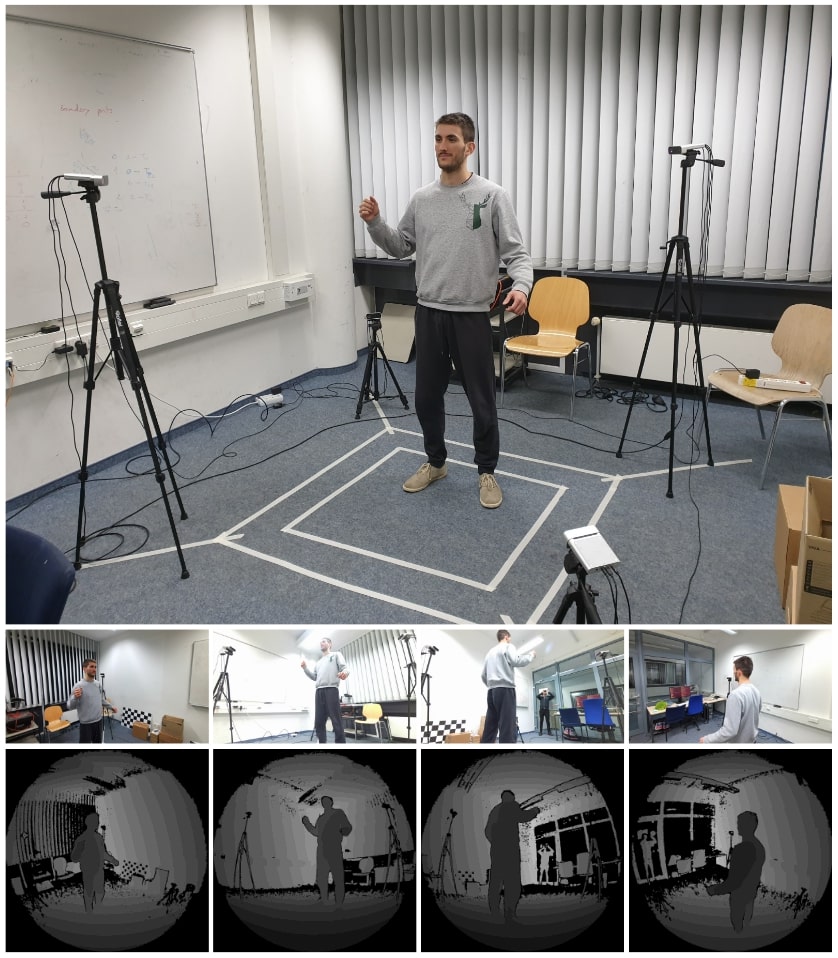}
    \caption{
    We capture non-rigid motion using 4 Kinect Azure sensors, which are pre-calibrated and hardware-synced to guarantee spatial and temporal coherence of depth captures.
    We capture depth images at resolution of $1024 \times 1024$, using the highest available frame-rate of $15$ FPS.
    }
    \label{fig:capture_setup}
\end{figure}

In the main paper as well as in the supplemental video, we demonstrate that our method performs well on real data.
To capture real-world non-rigid motion sequences, we record with four Kinect Azure sensors as shown in Fig.~\ref{fig:capture_setup}.
All four sensors are connected to the same computer via USB-C cables and are hardware-synced using a daisy-chain configuration (connecting the trigger of the master camera with the other 3 cameras in a chain).
To avoid interference between depth sensors, we set a delay of $160$ microseconds between different depth camera captures.

\end{appendix}

\newpage
{\small
\bibliographystyle{ieee_fullname}
\bibliography{main}

\begin{thebibliography}{10}\itemsep=-1pt

\bibitem{baran2007pinocchio}
Ilya Baran and Jovan Popovi{\'c}.
\newblock Automatic rigging and animation of 3d characters.
\newblock {\em ACM Transactions on graphics (TOG)}, 26(3):72--es, 2007.

\bibitem{bozic2020neuraltracking}
Alja{\v{z}} Bo{\v{z}}i{\v{c}}, Pablo Palafox, Michael Zoll{\"o}fer, Angela Dai,
  Justus Thies, and Matthias Nie{\ss}ner.
\newblock Neural non-rigid tracking.
\newblock In {\em NeurIPS}, 2020.

\bibitem{bozic2020deepdeform}
Alja{\v{z}} Bo{\v{z}}i{\v{c}}, Michael Zollh{\"o}fer, Christian Theobalt, and
  Matthias Nie{\ss}ner.
\newblock Deepdeform: Learning non-rigid rgb-d reconstruction with
  semi-supervised data.
\newblock In {\em Proceedings of the IEEE Conference on Computer Vision and
  Pattern Recognition (CVPR)}, 2020.

\bibitem{cao2018openpose}
Zhe Cao, Gines Hidalgo, Tomas Simon, Shih-En Wei, and Yaser Sheikh.
\newblock Openpose: realtime multi-person 2d pose estimation using part
  affinity fields.
\newblock {\em arXiv preprint arXiv:1812.08008}, 2018.

\bibitem{medioni1992}
Yang Chen and Gérard Medioni.
\newblock Object modeling by registration of multiple range images.
\newblock {\em Image Vision Comput.}, 10:145--155, 01 1992.

\bibitem{chibane2020ifnets}
Julian Chibane, Thiemo Alldieck, and Gerard Pons-Moll.
\newblock Implicit functions in feature space for 3d shape reconstruction and
  completion.
\newblock In {\em Proceedings of the IEEE/CVF Conference on Computer Vision and
  Pattern Recognition}, pages 6970--6981, 2020.

\bibitem{collet2015fvv}
Alvaro Collet, Ming Chuang, Pat Sweeney, Don Gillett, Dennis Evseev, David
  Calabrese, Hugues Hoppe, Adam Kirk, and Steve Sullivan.
\newblock High-quality streamable free-viewpoint video.
\newblock {\em ACM Transactions on Graphics (ToG)}, 34(4):1--13, 2015.

\bibitem{curless1996volumetric}
Brian Curless and Marc Levoy.
\newblock A volumetric method for building complex models from range images.
\newblock In {\em Proceedings of the 23rd annual conference on Computer
  graphics and interactive techniques}, pages 303--312, 1996.

\bibitem{deng2019nasa}
Boyang Deng, JP Lewis, Timothy Jeruzalski, Gerard Pons-Moll, Geoffrey Hinton,
  Mohammad Norouzi, and Andrea Tagliasacchi.
\newblock Neural articulated shape approximation.
\newblock In {\em The European Conference on Computer Vision (ECCV)}.
  {Springer}, August 2020.

\bibitem{dou2017motion2fusion}
Mingsong Dou, Philip Davidson, Sean~Ryan Fanello, Sameh Khamis, Adarsh Kowdle,
  Christoph Rhemann, Vladimir Tankovich, and Shahram Izadi.
\newblock Motion2fusion: Real-time volumetric performance capture.
\newblock {\em ACM Transactions on Graphics (TOG)}, 36(6):1--16, 2017.

\bibitem{dou20153d}
Mingsong Dou, Jonathan Taylor, Henry Fuchs, Andrew Fitzgibbon, and Shahram
  Izadi.
\newblock 3d scanning deformable objects with a single rgbd sensor.
\newblock In {\em Proceedings of the IEEE Conference on Computer Vision and
  Pattern Recognition (CVPR)}, pages 493--501, 2015.

\bibitem{genova2019sif}
Kyle Genova, Forrester Cole, Avneesh Sud, Aaron Sarna, and Thomas Funkhouser.
\newblock Deep structured implicit functions.
\newblock {\em arXiv preprint arXiv:1912.06126}, 2019.

\bibitem{genova2020ldif}
Kyle Genova, Forrester Cole, Avneesh Sud, Aaron Sarna, and Thomas Funkhouser.
\newblock Local deep implicit functions for 3d shape.
\newblock In {\em Proceedings of the IEEE/CVF Conference on Computer Vision and
  Pattern Recognition}, pages 4857--4866, 2020.

\bibitem{3DSemanticSegmentationWithSubmanifoldSparseConvNet}
Benjamin Graham, Martin Engelcke, and Laurens van~der Maaten.
\newblock 3d semantic segmentation with submanifold sparse convolutional
  networks.
\newblock {\em CVPR}, 2018.

\bibitem{guo2019relightables}
Kaiwen Guo, Peter Lincoln, Philip Davidson, Jay Busch, Xueming Yu, Matt Whalen,
  Geoff Harvey, Sergio Orts-Escolano, Rohit Pandey, Jason Dourgarian, et~al.
\newblock The relightables: Volumetric performance capture of humans with
  realistic relighting.
\newblock {\em ACM Transactions on Graphics (TOG)}, 38(6):1--19, 2019.

\bibitem{guo2017monofvv}
Kaiwen Guo, Feng Xu, Tao Yu, Xiaoyang Liu, Qionghai Dai, and Yebin Liu.
\newblock Real-time geometry, albedo, and motion reconstruction using a single
  rgb-d camera.
\newblock {\em ACM Transactions on Graphics (TOG)}, 36(3):32, 2017.

\bibitem{huang2020arch}
Zeng Huang, Yuanlu Xu, Christoph Lassner, Hao Li, and Tony Tung.
\newblock Arch: Animatable reconstruction of clothed humans.
\newblock In {\em Proceedings of the IEEE/CVF Conference on Computer Vision and
  Pattern Recognition}, pages 3093--3102, 2020.

\bibitem{innmann2020nrmvs}
Matthias Innmann, Kihwan Kim, Jinwei Gu, Matthias Nie{\ss}ner, Charles Loop,
  Marc Stamminger, and Jan Kautz.
\newblock Nrmvs: Non-rigid multi-view stereo.
\newblock In {\em The IEEE Winter Conference on Applications of Computer
  Vision}, pages 2754--2763, 2020.

\bibitem{innmann2016volumedeform}
Matthias Innmann, Michael Zollh{\"o}fer, Matthias Nie{\ss}ner, Christian
  Theobalt, and Marc Stamminger.
\newblock Volumedeform: Real-time volumetric non-rigid reconstruction.
\newblock In {\em European Conference on Computer Vision}, pages 362--379.
  Springer, 2016.

\bibitem{kazhdan2006poisson}
Michael Kazhdan, Matthew Bolitho, and Hugues Hoppe.
\newblock Poisson surface reconstruction.
\newblock In {\em Proceedings of the fourth Eurographics symposium on Geometry
  processing}, volume~7, 2006.

\bibitem{kazhdan2013screenedpoisson}
Michael Kazhdan and Hugues Hoppe.
\newblock Screened poisson surface reconstruction.
\newblock {\em ACM Transactions on Graphics (ToG)}, 32(3):1--13, 2013.

\bibitem{adam}
Diederik~P. Kingma and Jimmy Ba.
\newblock Adam: {A} method for stochastic optimization.
\newblock {\em CoRR}, abs/1412.6980, 2014.

\bibitem{lepetit2009epnp}
Vincent Lepetit, Francesc Moreno-Noguer, and Pascal Fua.
\newblock Epnp: An accurate o (n) solution to the pnp problem.
\newblock {\em International journal of computer vision}, 81(2):155, 2009.

\bibitem{li2020learning}
Yang Li, Aljaz Bozic, Tianwei Zhang, Yanli Ji, Tatsuya Harada, and Matthias
  Nie{\ss}ner.
\newblock Learning to optimize non-rigid tracking.
\newblock In {\em Proceedings of the IEEE/CVF Conference on Computer Vision and
  Pattern Recognition}, pages 4910--4918, 2020.

\bibitem{yang20204dcomplete}
Yang Li, Hikari Takehara, Takafumi Taketomi, Bo Zheng, and Matthias
  Nie{\ss}ner.
\newblock 4dcomplete: Non-rigid motion estimation beyond the observable
  surface.

\bibitem{loper2015smpl}
Matthew Loper, Naureen Mahmood, Javier Romero, Gerard Pons-Moll, and Michael~J
  Black.
\newblock Smpl: A skinned multi-person linear model.
\newblock {\em ACM transactions on graphics (TOG)}, 34(6):1--16, 2015.

\bibitem{marching_cubes}
William~E. Lorensen and Harvey~E. Cline.
\newblock Marching cubes: A high resolution 3d surface construction algorithm.
\newblock In {\em Proceedings of the 14th Annual Conference on Computer
  Graphics and Interactive Techniques}, SIGGRAPH '87, page 163–169, New York,
  NY, USA, 1987. Association for Computing Machinery.

\bibitem{mildenhall2020nerf}
Ben Mildenhall, Pratul~P Srinivasan, Matthew Tancik, Jonathan~T Barron, Ravi
  Ramamoorthi, and Ren Ng.
\newblock Nerf: Representing scenes as neural radiance fields for view
  synthesis.
\newblock {\em arXiv preprint arXiv:2003.08934}, 2020.

\bibitem{mueller2018ganeratedhands}
Franziska Mueller, Florian Bernard, Oleksandr Sotnychenko, Dushyant Mehta,
  Srinath Sridhar, Dan Casas, and Christian Theobalt.
\newblock Ganerated hands for real-time 3d hand tracking from monocular rgb.
\newblock In {\em Proceedings of the IEEE Conference on Computer Vision and
  Pattern Recognition}, pages 49--59, 2018.

\bibitem{newcombe2015dynamicfusion}
Richard~A Newcombe, Dieter Fox, and Steven~M Seitz.
\newblock Dynamicfusion: Reconstruction and tracking of non-rigid scenes in
  real-time.
\newblock In {\em Proceedings of the IEEE Conference on Computer Vision and
  Pattern Recognition (CVPR)}, pages 343--352, 2015.

\bibitem{park2019deepsdf}
Jeong~Joon Park, Peter Florence, Julian Straub, Richard Newcombe, and Steven
  Lovegrove.
\newblock Deepsdf: Learning continuous signed distance functions for shape
  representation.
\newblock In {\em Proceedings of the IEEE Conference on Computer Vision and
  Pattern Recognition}, pages 165--174, 2019.

\bibitem{sidhu2020neuralnonrigidsfm}
Vikramjit Sidhu, Edgar Tretschk, Vladislav Golyanik, Antonio Agudo, and
  Christian Theobalt.
\newblock Neural dense non-rigid structure from motion with latent space
  constraints.
\newblock In {\em European Conference on Computer Vision (ECCV)}, 2020.

\bibitem{sitzmann2019scene}
Vincent Sitzmann, Michael Zollh{\"o}fer, and Gordon Wetzstein.
\newblock Scene representation networks: Continuous 3d-structure-aware neural
  scene representations.
\newblock In {\em Advances in Neural Information Processing Systems}, pages
  1121--1132, 2019.

\bibitem{slavcheva2017killingfusion}
Miroslava Slavcheva, Maximilian Baust, Daniel Cremers, and Slobodan Ilic.
\newblock Killingfusion: Non-rigid 3d reconstruction without correspondences.
\newblock In {\em Proceedings of the IEEE Conference on Computer Vision and
  Pattern Recognition (CVPR)}, pages 1386--1395, 2017.

\bibitem{slavcheva2018sobolevfusion}
Miroslava Slavcheva, Maximilian Baust, and Slobodan Ilic.
\newblock Sobolevfusion: 3d reconstruction of scenes undergoing free non-rigid
  motion.
\newblock In {\em Proceedings of the IEEE Conference on Computer Vision and
  Pattern Recognition (CVPR)}, pages 2646--2655, 2018.

\bibitem{sorkine2007arap}
Olga Sorkine and Marc Alexa.
\newblock As-rigid-as-possible surface modeling.
\newblock In {\em Symposium on Geometry processing}, volume~4, pages 109--116,
  2007.

\bibitem{sumner2007ed}
Robert~W Sumner, Johannes Schmid, and Mark Pauly.
\newblock Embedded deformation for shape manipulation.
\newblock In {\em ACM SIGGRAPH 2007 papers}, pages 80--es. 2007.

\bibitem{thies2016face2face}
Justus Thies, Michael Zollhofer, Marc Stamminger, Christian Theobalt, and
  Matthias Nie{\ss}ner.
\newblock Face2face: Real-time face capture and reenactment of rgb videos.
\newblock In {\em Proceedings of the IEEE conference on computer vision and
  pattern recognition}, pages 2387--2395, 2016.

\bibitem{tretschk2020patchnets}
Edgar Tretschk, Ayush Tewari, Vladislav Golyanik, Michael Zollh{\"o}fer,
  Carsten Stoll, and Christian Theobalt.
\newblock Patchnets: Patch-based generalizable deep implicit 3d shape
  representations.
\newblock In {\em European Conference on Computer Vision}, pages 293--309.
  Springer, 2020.

\bibitem{tretschk2019demea}
Edgar Tretschk, Ayush Tewari, Michael Zollh{\"o}fer, Vladislav Golyanik, and
  Christian Theobalt.
\newblock Demea: Deep mesh autoencoders for non-rigidly deforming objects.
\newblock {\em arXiv preprint arXiv:1905.10290}, 2019.

\bibitem{wang2018dynamic}
Sen Wang, Xinxin Zuo, Chao Du, Runxiao Wang, Jiangbin Zheng, and Ruigang Yang.
\newblock Dynamic non-rigid objects reconstruction with a single rgb-d sensor.
\newblock {\em Sensors}, 18(3):886, 2018.

\bibitem{xu2020rignet}
Zhan Xu, Yang Zhou, Evangelos Kalogerakis, Chris Landreth, and Karan Singh.
\newblock Rignet: Neural rigging for articulated characters.
\newblock {\em arXiv preprint arXiv:2005.00559}, 2020.

\end{thebibliography}
}

\end{document}